\documentclass[preprint,12pt]{elsarticle}
\usepackage{tabularx}
\usepackage{graphicx}
\usepackage{adjustbox}
\usepackage{lineno}
\usepackage{setspace}
\usepackage{amsmath}
\usepackage{booktabs}
\usepackage{ltablex}
\usepackage{pdflscape}
\usepackage{hyperref}

\usepackage{amssymb}

\begin{document}

\begin{frontmatter}


\doublespacing

\title{Mini-JEPA Foundation Model Fleet Enables Agentic Hydrologic Intelligence}

\author[inst1]{Mashrekur Rahman}
\ead{mashrekur.rahman@dartmouth.edu}

\affiliation[inst1]{organization={Dartmouth Libraries, Dartmouth College},
            addressline={6025 Baker-Berry Library}, 
            city={Hanover},
            postcode={03755}, 
            state={NH},
            country={USA}}

\begin{abstract}
Geospatial foundation models leverage multispectral observations into dense embeddings that are increasingly applied in natural-language environmental reasoning systems. A single planetary-scale model, e.g. Google AlphaEarth, handles broad characterization well but may compromise on specialized hydrologic signals. Such generalist models are also often inaccessible, expensive, and require large-scale compute. We propose a fleet of small sensor-specialized Joint Embedding Predictive Architecture (JEPA) foundation models that can be consulted by a routing agent for specialized questions. We pretrained five (22M parameter) \textbf{Mini-JEPAs} that share an identical Vision Transformer backbone, an identical JEPA training recipe, and a 64-dimensional output space, using images from Sentinel-2 optical, Sentinel-1 SAR, MODIS thermal, multi-temporal Sentinel-2 phenology, and a topography and soil stack. Each Mini-JEPA reconstructs the environmental variable matched to its sensor, with cross-validated $R^2$ reaching 0.97 for elevation under the topography-soil model, 0.97 for mean temperature under the thermal model, and 0.81 for precipitation under the phenology model. Across all five models, the variable each one predicts best is the variable its sensor physically observes. The five manifolds differ in geometric structure, with global participation ratios ranging from 8.9 for the SAR model to 20.2 for the thermal model, and local intrinsic dimensionalities from 2.3 to 9.0. For contrast, the joint topography-soil and phenology models add predictive value beyond AlphaEarth Foundation Model alone for soil moisture, aridity, and precipitation ($\Delta R^2$ up to 0.031). A router LLM (Claude Sonnet/Opus) reads per-modality references and selects the appropriate sensors with a perfect expected-modality hit rate over a curated question set. In paired LLM-as-Judge evaluation, dual retrieval over AlphaEarth and the routed fleet outperforms AlphaEarth alone on questions whose signal matches a single sensor (Cohen's $d = 1.10$, $p = 0.031$). Aggregate gains across all question types remain modest. On the physics-matched questions, the routed fleet alone also scores comparably to AlphaEarth. Our findings suggest that locally-trained specialized Mini-JEPAs can be operationalized for hydrologic intelligence systems with modest compute, providing a complement or substitute to planetary-scale foundation models.
\end{abstract}

\begin{highlights}
\item Mini-JEPAs match planetary-scale generalists on physics-matched tasks
\item Each Mini-JEPA captures the physical signal of its satellite sensor
\item The five Mini-JEPAs yield distinct embedding manifold geometries
\item The fleet is operationalized for agentic hydrologic intelligence
\end{highlights}


\end{frontmatter}

\doublespacing

\section{Introduction} \label{sec:introduction}

Satellite foundation models have become a popular paradigm for processing global earth observation data into reusable embedding representations \citep{brown2025alphaearth, bodnar2025foundation, jakubik2023prithvi, xiao2025foundation}. These models compress multispectral, radar, and atmospheric observations into dense vectors that downstream systems use for retrieval, classification, and natural-language reasoning over the land surface \citep{mai2023opportunities, rahman2026physically, rahman2026characterizing}. The field has generally gravitated towards a common practice: train a large model on as much data as possible, then expose its embeddings to downstream applications through retrieval interfaces.

This practice has limits that grow more apparent as the downstream questions get more specific. A single embedding space must allocate its dimensions across phenomena whose physics live in distinctly different sensors. Optical reflectance, microwave backscatter, thermal emission, and seasonal phenology each tell us different stories about the surface, and a generalist embedding may compromise across them. The compromise is usually not apparent for broad characterization tasks and conspicuous for specialized hydrologic ones, where the relevant signal may sit in a sensor the generalist  can underweigh. There is also a practical caveat - planetary-scale models are expensive to train, frequently closed, and slow to extend or audit. Research groups and institutions without heavy compute resources or costly API access cannot easily build, retrain, or operationalize on their own - making the space for alternative foundation modeling paradigms.

Joint Embedding Predictive Architecture or JEPA \citep{lecun2022path, assran2023ijepa, bardes2022vicreg} trains an encoder to predict masked latent representations rather than predicting pixel values. Predicting in latent space frees the objective from the input modality: the same principle applies to optical imagery, SAR backscatter, thermal observations, etc., with no reconstruction-loss bias toward one sensor's noise profile. This matters here because if the architecture and the recipe are held fixed, any structural differences between models trained on different sensors must come from the sensors themselves rather than from training choices. JEPA also avoids the computational cost of pixel-level reconstruction, since prediction targets are low-dimensional latents rather than full image patches, which reduces pretraining compute compared to masked autoencoder approaches \citep{he2022mae, assran2023ijepa}.

In this study, we propose \textbf{Mini-JEPAs}: a fleet of small sensor-specialized foundation models for hydrologic intelligence, paired with a routing agent that decides which specialists to consult for a given question. Each Mini-JEPA shares an identical Vision Transformer backbone \citep{dosovitskiy2021vit}, identical JEPA training recipe, and a 64-dimensional output space, differing only in the satellite sensor product it was trained on. In previous studies, we first \citep{rahman2026physically} established that individual dimensions of a planetary-scale embedding space (AlphaEarth) encode physically meaningful properties of the land surface. Secondly, \citep{rahman2026characterizing} characterized the geometric structure of that same embedding space and built an agentic system that uses it as the retrieval substrate. The present work extends both methods to a multi-model setting, asking what changes when the substrate is a fleet of small specialists rather than one generalist.

We ask the following specific research questions:

\begin{enumerate}
    \item Does each Mini-JEPA reconstruct the environmental variable physically matched to its sensor, and is the specialization clean across the fleet?
    \item Do the Mini-JEPA models produce embedding manifolds with distinct geometric structures, and do those structures reflect the physics of their respective sensors?
    \item Do Mini-JEPAs carry information beyond a planetary-scale generalist, and on which variables is the complementarity strongest?
    \item Can a router LLM compose the fleet into an agentic system whose retrievals enable hydrologic reasoning?
\end{enumerate}

\section{Methods} \label{sec:methods}

\subsection{Data} \label{sec:methods_data}

\subsubsection{Patch grid}

We sampled 10,000 candidate patch centers uniformly at random across the Continental United States (CONUS), seeded with the same random state used in \citet{rahman2026physically, rahman2026characterizing}. Each patch covered a $128 \times 128$ pixel window at 30~m ground sample distance, corresponding to a $3.84~\text{km} \times 3.84~\text{km}$ footprint on the surface. After excluding water bodies, mostly-cloud-contaminated scenes, and patches with incomplete sensor coverage, 9{,}704 patches remained as our working corpus. The same 9,704 patch centers were reused across all five satellite modalities, so that any downstream differences between the Mini-JEPAs reflect what each sensor saw at those locations rather than differences in geographic sampling. All images were retrieved from Google Earth Engine \citep{gorelick2017gee} for the calendar year 2022. A single calendar year was sufficient to demonstrate the fleet design, and kept pretraining tractable on a single workstation GPU.

\begin{figure}[t]
\centering
\includegraphics[width=1.1\linewidth]{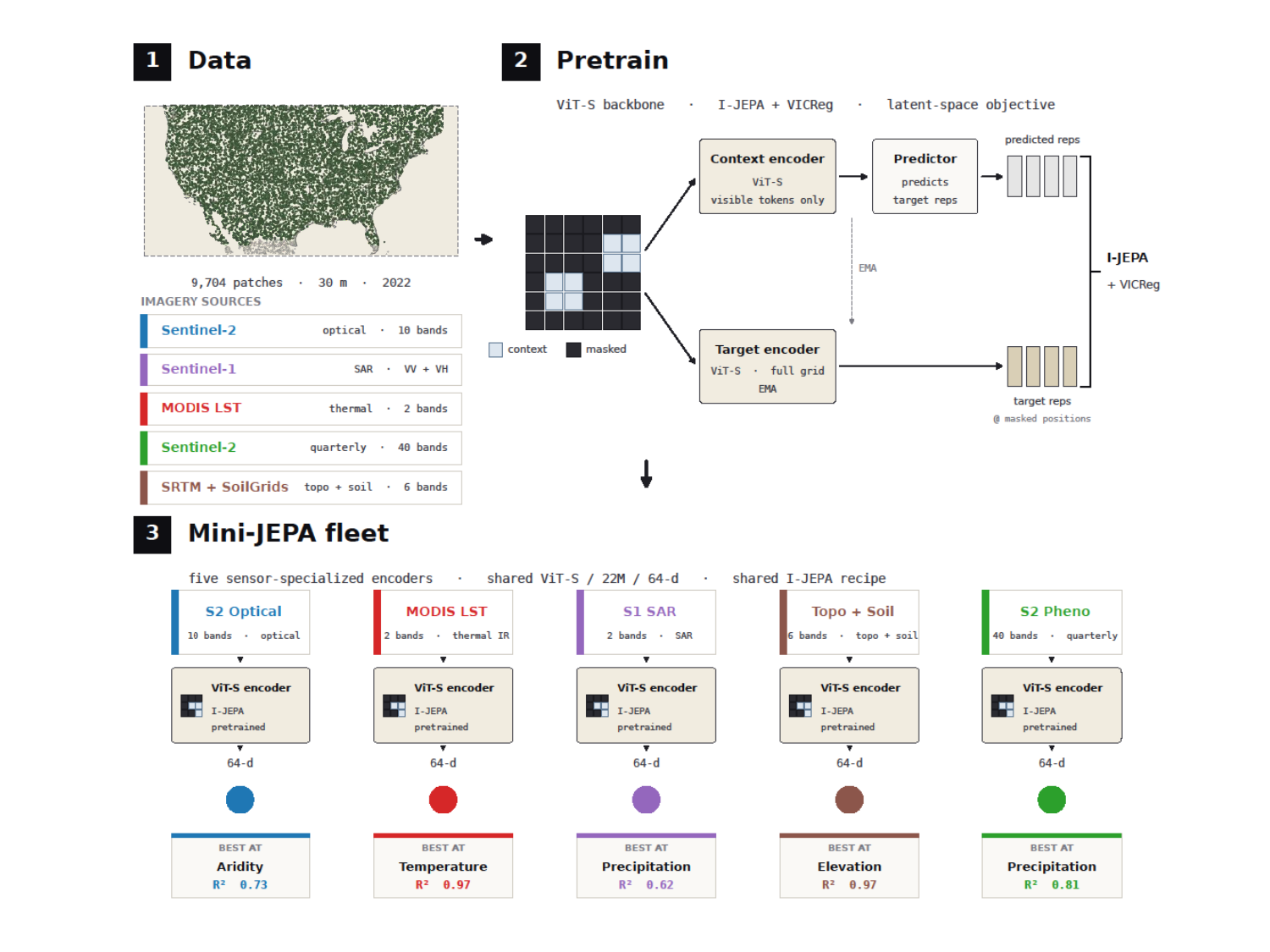}
\caption{Overview of the Mini-JEPA fleet. \textbf{Panel 1 (Data):} 9,704 patch centers sampled across CONUS at 30~m resolution, paired with imagery from five Google Earth Engine sensor products (Sentinel-2 optical, Sentinel-1 SAR, MODIS thermal, multi-temporal Sentinel-2 phenology, SRTM + SoilGrids topography and soil). \textbf{Panel 2 (Pretrain):} I-JEPA + VICReg pretraining loop. The context encoder sees visible tokens, the target encoder (an exponential moving average) sees the full grid, and a predictor learns to map context to target latents at masked positions. \textbf{Panel 3 (Mini-JEPA fleet):} the five resulting sensor-specialized encoders. All share an identical ViT-S backbone (22M parameters, 64-d output) and identical I-JEPA recipe; they differ only in the sensor product used during pretraining. Color coding for each modality is preserved across all subsequent figures.}
\label{fig:overview}
\end{figure}

\subsubsection{Satellite products}

Each Mini-JEPA was trained on one of five Google Earth Engine satellite products, chosen to span complementary aspects of land surface physics. Sentinel-2 surface reflectance (annual median, ten optical and shortwave-infrared bands) gives broad spectral information about vegetation, soil, and surface composition. Sentinel-1 dual-polarization SAR (VV+VH, annual median) sees through clouds and is sensitive to surface roughness, soil moisture, and inundation. MODIS land surface temperature (day and night composites) measures thermal emission, capturing evaporative cooling, urban heat islands, and frost regime. Sentinel-2 phenology (four quarterly composites, ten bands each) captures seasonal vegetation dynamics that the annual median averages away. Finally, a topography and soil stack combines SRTM elevation, slope, and aspect with three SoilGrids surface-soil properties \citep{poggio2021soilgrids, farr2007srtm}. Detailed source IDs, band counts, and physical interpretations are listed in Table~\ref{tab:datasets}. Figure~\ref{fig:overview}, panel 1, shows the spatial coverage and inputs.

\subsubsection{Environmental labels}

For per-model evaluation we paired each patch with co-located environmental variables: SMAP surface soil moisture \citep{entekhabi2010smap}, PRISM mean annual precipitation and temperature \citep{daly2008prism}, NLCD land cover class \citep{dewitz2021nlcd}, K\"oppen-Geiger climate class \citep{beck2018koppen}, SRTM elevation, and an aridity proxy ($P/\text{PET}$). For complementarity analysis, we also retrieved the 64-dimensional AlphaEarth embedding at each patch center \citep{brown2025alphaearth}. The labels were not used during pretraining; they only entered the per-model evaluation pipeline (Section~\ref{sec:methods_evaluation}).

\begin{table}[!t]
\centering
\caption{Satellite products used to train each Mini-JEPA. All five products share the same 9{,}704 patch locations at 30~m resolution. Modality colors match Figure~\ref{fig:overview} and all subsequent figures.}
\label{tab:datasets}
\footnotesize
\setlength{\tabcolsep}{4pt}
\renewcommand{\arraystretch}{1.15}
\begin{tabularx}{\linewidth}{@{} l l c >{\raggedright\arraybackslash}X @{}}
\toprule
\textbf{Mini-JEPA} & \textbf{Source} & \textbf{Bands} & \textbf{Physical signal} \tabularnewline
\midrule
S2-Optical    & Sentinel-2 SR (annual median) \citep{drusch2012sentinel2} & 10 & Reflectance over visible, red-edge, NIR, and SWIR bands \tabularnewline
\addlinespace
S1-SAR        & Sentinel-1 GRD (VV+VH median) \citep{torres2012sentinel1} & 2  & Microwave backscatter; sees through clouds; sensitive to roughness, soil moisture, inundation \tabularnewline
\addlinespace
MODIS-Thermal & MODIS LST (day + night) \citep{wan2014modis}              & 2  & Thermal emission; evaporative cooling, urban heat islands, frost regime \tabularnewline
\addlinespace
S2-Phenology  & Sentinel-2 SR (four quarterly composites)                 & 40 & Seasonal vegetation dynamics; agriculture and deciduous-evergreen contrast \tabularnewline
\addlinespace
Topo-Soil     & SRTM + SoilGrids \citep{farr2007srtm, poggio2021soilgrids} & 6  & Elevation, slope, aspect, and surface soil properties \tabularnewline
\bottomrule
\end{tabularx}
\end{table}

\subsection{Mini-JEPA pretraining} \label{sec:methods_pretraining}

\subsubsection{Why JEPA, in practical terms}

Self-supervised pretraining gives a foundation model its initial sense of what each input looks like, without needing labels. There are two broad families of approaches. Masked autoencoders (MAE) \citep{he2022mae} hide part of the image and ask the model to reconstruct the missing pixels. And Joint Embedding Predictive Architectures (JEPA) \citep{lecun2022path, assran2023ijepa} hide part of the image and ask the model to predict what its \emph{representation} would have been, in the model's own learned latent space. The difference matters for our settings because pixel reconstruction forces the model to learn the noise statistics of each sensor; even the speckle in SAR or the striping artifacts in some MODIS scenes, even though those details have no value for downstream applications. Latent prediction sidesteps this cost: the model only spends capacity on features that predict latent structure, not pixels. The same idea can then be applied to optical imagery, SAR backscatter, thermal observations, or temporal phenology stacks without retuning the loss to any one sensor's noise profile.

\subsubsection{Shared architecture and recipe}

Each Mini-JEPA uses an identical Vision Transformer-Small (ViT-S) backbone \citep{dosovitskiy2021vit} with 12 transformer layers, 6 attention heads, and a hidden dimension of 384. Each $128 \times 128$ image is tokenized into 64 patches of $16 \times 16$ pixels. The final encoder representation is linearly projected to a 64-dimensional output space, matching AlphaEarth's dimensionality so that the two are directly comparable in downstream retrieval and complementarity analyses. The training recipe combines I-JEPA \citep{assran2023ijepa} with a VICReg regularizer \citep{bardes2022vicreg} that prevents representation collapse. The context encoder sees 60\% of the tokens; the target encoder, an exponential moving average of the context encoder, sees all tokens; a small predictor network learns to map context to target latents at masked positions. Each Mini-JEPA was trained for 100 epochs with batch size 64, learning rate $1.5 \times 10^{-4}$. Pretraining took a few hours per Mini-JEPA on a single NVIDIA RTX 5090, which makes the fleet easy to extend or retrain.

The five Mini-JEPAs share every hyperparameter listed above. The only difference between them is the satellite product the encoder sees during pretraining. Figure~\ref{fig:overview}, panels 2 and 3, shows the I-JEPA pretraining loop and the resulting fleet.

\subsection{Per-modality evaluation} \label{sec:methods_evaluation}

We characterized each Mini-JEPA along three axes that mirror the methodology established in our previous studies on AlphaEarth \citep{rahman2026physically, rahman2026characterizing}. Applying the same three-axis characterization to a fleet of specialists, rather than to a single generalist, lets us ask whether sensor specialization shows up consistently across all three views.

\subsubsection{Dimension-level physical interpretability}

For each Mini-JEPA we asked which of its 64 embedding dimensions encode which physical variables. We computed Spearman rank correlations between every embedding dimension and every environmental variable, then trained Random Forest regressors that map the full 64-d embedding to each variable and recorded permutation importance per dimension \citep{breiman2001randomforest}. To guard against spatial autocorrelation inflating naive cross-validation scores \citep{ploton2020spatial, roberts2017cross}, we used spatial-block cross-validation in which the training and test patches in each fold come from non-overlapping geographic blocks. 

\subsubsection{Manifold geometry}

For each Mini-JEPA, we characterized the geometric structure of its 64-d embedding space. We measured global participation ratio (PR), defined from the eigenvalue spectrum of the embedding covariance matrix, as a continuous measure of effective dimensionality \citep{gao2017theory}. We also estimated local intrinsic dimensionality via maximum likelihood on $k$-nearest-neighbor distances \citep{levina2004mle}, and ran local principal component analysis at each of 2{,}000 probe points to measure how many dimensions are active in a local neighborhood (local $n_{80}$, the number of PCs needed to capture 80\% of local variance). These three measures together describe how the embedding fills its 64-d space at a global scale, at the scale of an individual neighborhood, and at intermediate scales.

\subsubsection{Complementarity with AlphaEarth}

To test whether each Mini-JEPA carries information that AlphaEarth does not already represent, we fit canonical correlation analysis (CCA) between each Mini-JEPA's embeddings and AlphaEarth's embeddings at the same patch locations \citep{hotelling1936cca}. Low canonical correlations across many components indicate the two spaces are largely complementary; high correlations on the leading components indicate redundancy. We also fit Random Forest regressors that predict each environmental variable from three different feature sets: AlphaEarth alone, the Mini-JEPA alone, and the two concatenated. The joint predictive gain is the difference between the joint $R^2$ and the better of the two single-source $R^2$ values, with positive gain indicating that combining the two sources adds information that neither one carries on its own.

\subsection{Agentic system} \label{sec:methods_agent}

\subsubsection{Per-modality reference cards}

For each Mini-JEPA, we compiled a compact reference card summarizing what that model is good for. Each card contains the model's dimension dictionary (which dimensions encode which environmental variables, derived from the interpretability analysis above), its geometric profile (global PR, intrinsic dimensionality, mean local PR), a one-sentence statement of the sensor's physical signal and what it cannot see (for instance, S2-Optical cannot see through clouds; S1-SAR can), and a small table of per-variable cross-validated $R^2$ scores. The cards are short by design — they have to fit inside the router LLM's prompt and each summarizes the empirical findings of the per-modality evaluation in a form the agent can read.

\subsubsection{Routing and retrieval}

When a user submits a natural-language hydrologic question with optional geographic context, the router LLM (Claude Sonnet 4.6 or Opus 4.7) reads the per-modality reference cards and emits a structured tool-call plan. The plan specifies which Mini-JEPAs to query (a subset of the five, in any combination) and, optionally, whether to additionally query AlphaEarth. Each selected retrieval tool runs a $k$-nearest-neighbor search against a per-modality FAISS index \citep{johnson2019faiss} built from the mean-pooled 64-d embeddings of all 9{,}704 patches. The query vector is the embedding of a context patch encoded by the corresponding Mini-JEPA. A synthesis LLM then consumes the ranked retrievals from all selected modalities, with explicit modality provenance, and composes a final answer. The full agent flow is shown in Figure~\ref{fig:agent_architecture}.

\subsection{LLM-as-Judge evaluation} \label{sec:methods_evaluation_llm}

\subsubsection{Question set}

We constructed a curated set of 40 hydrologic questions stratified into four categories chosen to stress different parts of the fleet. The \emph{single-modality} category contains questions whose physical signal sits cleanly in one sensor (for instance, ``What is the dominant land cover near $43.4^\circ$N, $93.5^\circ$W?''). The \emph{multi-modality} category contains questions that require combining signals from multiple sensors. The \emph{SAR-favorable} category contains questions about conditions that defeat optical imagery (cloud cover, surface water under vegetation). The \emph{AE-favorable} category contains broad characterization questions where a planetary-scale generalist is expected to do well. Each question is paired with an expected-modality set used to evaluate routing accuracy.

\subsubsection{Conditions and scoring}

For each question we generated responses under three conditions: \textbf{AE only} (retrieval restricted to AlphaEarth), \textbf{Fleet only} (router selects from the five Mini-JEPAs, AlphaEarth disabled), and \textbf{AE + Fleet} (router can select from all six, the full dual-RAG setup). Each response was scored by two LLM judges (Claude Haiku 4.5 and GPT-OSS-120B) on five rubric items (grounding, scientific accuracy, completeness, coherence, practical utility), combined into a single weighted score on a 1-to-5 scale \citep{zheng2023judging}. The two judges were applied independently to every response, and effect sizes between conditions are reported as Cohen's $d$ with paired bootstrap $p$-values. Inter-judge calibration was assessed by comparing per-question $\Delta$ scores across judges.

\section{Results} \label{sec:results}

We present results in four parts, following the research questions posed in Section~\ref{sec:introduction}. Section~\ref{sec:results_skill} examines whether each Mini-JEPA reconstructs the environmental variable matched to its sensor. Section~\ref{sec:results_geometry} examines whether the five fleet members produce embedding manifolds with distinct geometric structures. Section~\ref{sec:results_complementarity} examines whether the Mini-JEPAs add information beyond AlphaEarth. Section~\ref{sec:results_agent} reports the routing accuracy and the LLM-as-Judge evaluation of the full agentic system.

\subsection{Sensor physics drives per-modality predictive skill} \label{sec:results_skill}

\begin{figure}[t]
\centering
\includegraphics[width=\linewidth]{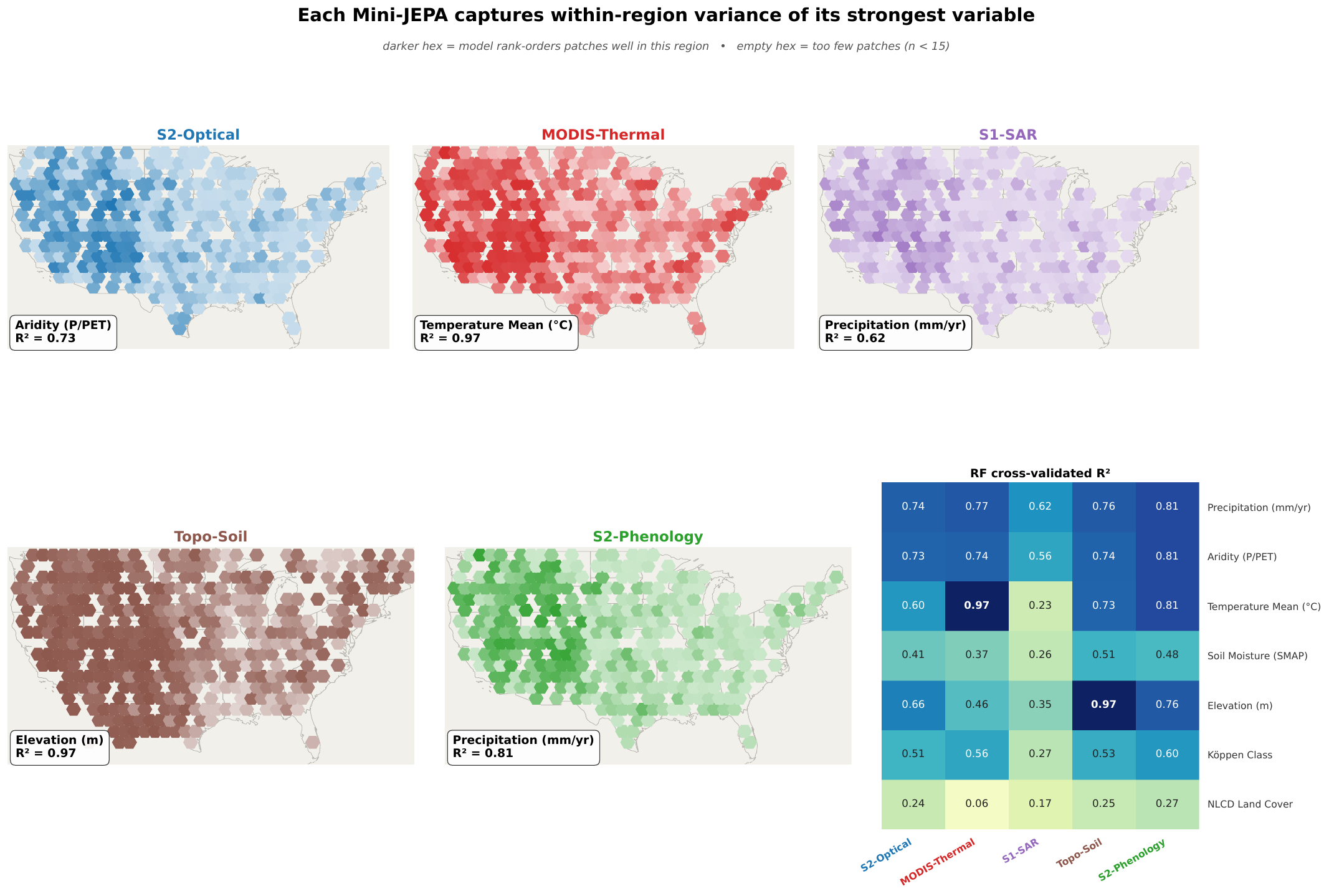}
\caption{Per-modality predictive skill across seven environmental variables. \textbf{Top and middle rows:} hex maps showing where each Mini-JEPA's predictions are accurate (darker = higher within-region $R^2$); the value shown in the inset is each model's strongest variable. \textbf{Bottom right:} cross-validated $R^2$ heatmap (rows = environmental variable, columns = Mini-JEPA). The strongest score in each row is bolded. The diagonal pattern, where each Mini-JEPA's best variable matches its sensor physics, is the main finding of this section.}
\label{fig:per_modality_skill}
\end{figure}

The first question is whether each Mini-JEPA reconstructs the environmental variable physically matched to its sensor. Figure~\ref{fig:per_modality_skill} reports cross-validated $R^2$ for seven environmental variables across all five Mini-JEPAs.

Each Mini-JEPA's strongest variable is the one its sensor physically observes. S2-Optical, which sees broad-band surface reflectance, predicts aridity best ($R^2 = 0.73$); aridity is closely tied to vegetation cover and soil exposure, both of which are encoded in optical reflectance. S1-SAR, which measures microwave backscatter and is sensitive to surface roughness and moisture, predicts mean annual precipitation best ($R^2 = 0.62$). MODIS-Thermal reaches its peak on mean annual temperature ($R^2 = 0.97$), the variable it most directly observes. S2-Phenology, which encodes seasonal vegetation dynamics, predicts precipitation best ($R^2 = 0.81$), and also reaches comparable scores on aridity and mean temperature. Topo-Soil reaches its peak on elevation ($R^2 = 0.97$), matching the input it most directly contains. The diagonal is visible in the heatmap of Figure~\ref{fig:per_modality_skill}: each Mini-JEPA's strongest score sits where its sensor physics predicts.

The hex maps in the upper panels of Figure~\ref{fig:per_modality_skill} show where each Mini-JEPA's predictive skill is geographically expressed. Darker hexes indicate regions where the model rank-orders patches well in within-region cross-validation. The MODIS-Thermal hex map is darkest across the climatologically variable interior, where temperature gradients dominate. The Topo-Soil map is darkest in the topographically complex West, where elevation varies sharply over short distances. S2-Phenology shows a green diagonal across the agricultural Midwest and the Great Plains, where seasonal vegetation dynamics are most pronounced. The geographic patterns suggest that each Mini-JEPA's skill is not uniform but is concentrated in regions where the physical phenomenon it sees is itself spatially expressed.

Two further observations are worth noting. First, the leading score across the fleet for any given variable is rarely produced by the sensor we would have nominated naively. Mean annual precipitation, for instance, peaks not under the SAR model (which is moisture-sensitive) but under S2-Phenology (which sees the vegetation response to precipitation). The fleet's pattern of strengths is therefore not a simple sensor-to-variable lookup; it reflects how each sensor's physical signal maps onto the downstream variable through the surface processes between them. Second, the worst-predicted variable for every Mini-JEPA is NLCD land cover, a categorical label that none of the continuous sensor signals decodes cleanly at 30~m resolution from a 22M-parameter encoder. The relative weakness on NLCD is shared across the fleet, suggesting it is a property of the task rather than of any particular Mini-JEPA.

\subsection{The five Mini-JEPAs produce distinct manifold geometries} \label{sec:results_geometry}

\begin{figure}[t]
\centering
\includegraphics[width=\linewidth]{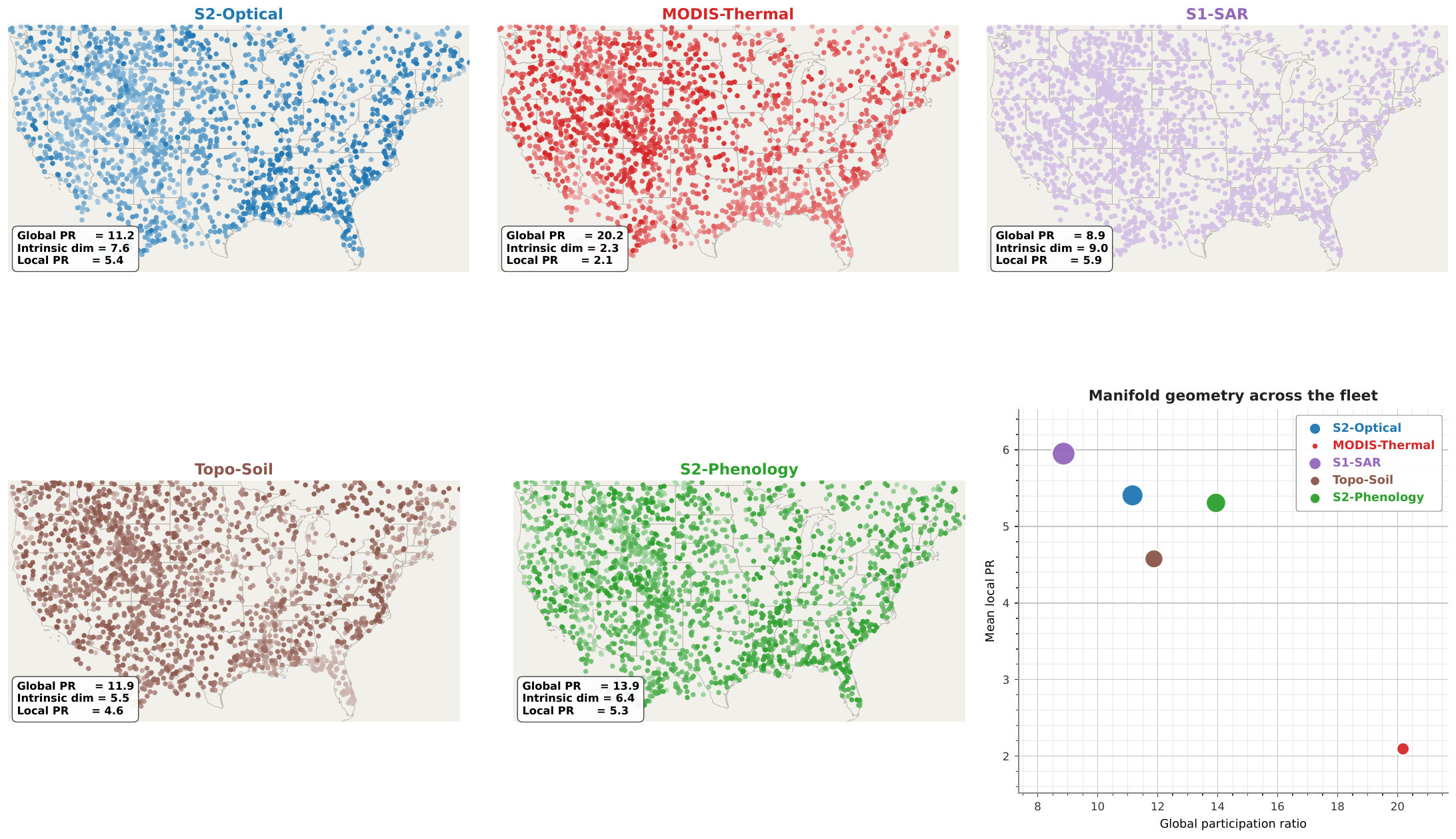}
\caption{Manifold geometry across the Mini-JEPA fleet. Per-modality maps show the spatial distribution of 2{,}000 local PCA probes; insets report global participation ratio, intrinsic dimensionality, and mean local PR. The scatter at bottom right plots mean local PR against global PR for all five models. MODIS-Thermal sits in the high-global, low-local corner (one-dimensional locally); S1-SAR sits in the opposite corner.}
\label{fig:manifold_geometry}
\end{figure}

\begin{figure}[t]
\centering
\includegraphics[width=\linewidth]{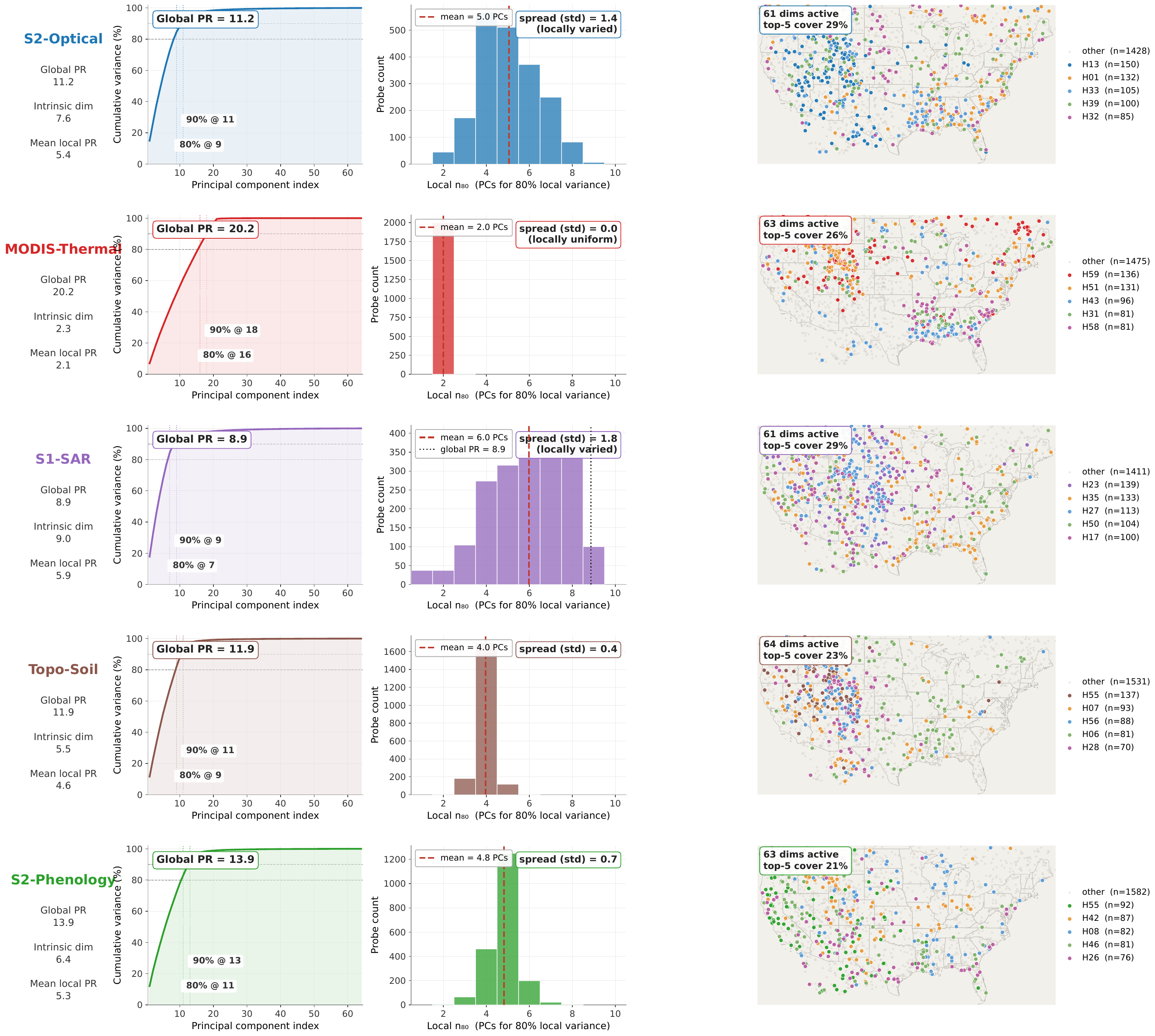}
\caption{Manifold portrait for each Mini-JEPA. \textbf{Left column:} cumulative variance spectrum with the principal-component indices needed to reach 80\% and 90\% of total variance. \textbf{Middle column:} histogram of local $n_{80}$ values across 2{,}000 probe points; lower spread indicates a more spatially uniform local geometry. \textbf{Right column:} CONUS maps of the locally-dominant embedding dimension at each probe; legend lists the top five dimensions and the fraction of probes they cover.}
\label{fig:manifold_portrait}
\end{figure}

The second question is whether the embedding spaces of the five Mini-JEPAs are geometrically distinct, and whether those differences reflect sensor physics. Figures~\ref{fig:manifold_geometry} and~\ref{fig:manifold_portrait} characterize the five manifolds.

The global participation ratio (PR), computed from the eigenvalue spectrum of each Mini-JEPA's embedding covariance, varies widely across the fleet (Figure~\ref{fig:manifold_geometry}). S1-SAR has the lowest global PR (8.9), indicating that its 64-d embedding lives close to a relatively low-dimensional global subspace. S2-Optical and Topo-Soil sit near 11-12. S2-Phenology reaches 13.9. MODIS-Thermal sits at the higher end with a global PR of 20.2. The five PR values span more than a factor of two, even though the encoders are architecturally identical.

The local picture is roughly the inverse. Maximum-likelihood intrinsic dimensionality and the local $n_{80}$ statistic (how many principal components are needed to capture 80\% of variance in a local neighborhood) both show that MODIS-Thermal's manifold is near one-dimensional locally (mean local $n_{80} = 2.0$, std $= 0.0$), while S1-SAR's manifold is locally the most complex (mean local $n_{80} = 6.0$, std $= 1.8$). S2-Optical, Topo-Soil, and S2-Phenology sit in between (mean local $n_{80} \approx 4-5$).

The MODIS-Thermal and S1-SAR patterns are physically consistent with what each sensor measures. Land surface temperature varies smoothly along a single dominant gradient (latitude crossed with elevation), so a thermal encoder trained on a year of MODIS LST sees most of the variance along one axis. The high global PR reflects the encoder using many dimensions to express this gradient at different geographic locations, but the local picture is one-dimensional because any small neighborhood lies on a near-linear stretch of the gradient. SAR backscatter, in contrast, responds simultaneously to surface roughness, soil moisture, biomass, and inundation, and these signals decorrelate at local scales. The S1-SAR manifold therefore reaches its complexity locally rather than globally.

The right-hand panels of Figure~\ref{fig:manifold_portrait} show, for each Mini-JEPA, which embedding dimension is locally dominant at each of the 2,000 probe points across CONUS. The top-5 dominant dimensions account for 21\% (S2-Phenology) to 29\% (S2-Optical, S1-SAR) of probes, indicating that no Mini-JEPA reduces to a small set of dominant directions, and that all five use most of their available dimensions to express local variation. MODIS-Thermal is the partial exception: while it has 63 active dimensions globally, locally it concentrates on a single dimension everywhere, consistent with the locally-uniform $n_{80}$ histogram.

The five Mini-JEPAs therefore do not share a manifold structure. They differ in both their global effective dimensionality and their local geometric complexity, and the differences track what each sensor physically observes.

\subsection{Mini-JEPAs complement AlphaEarth on hydrologic variables} \label{sec:results_complementarity}

\begin{figure}[t]
\centering
\includegraphics[width=\linewidth]{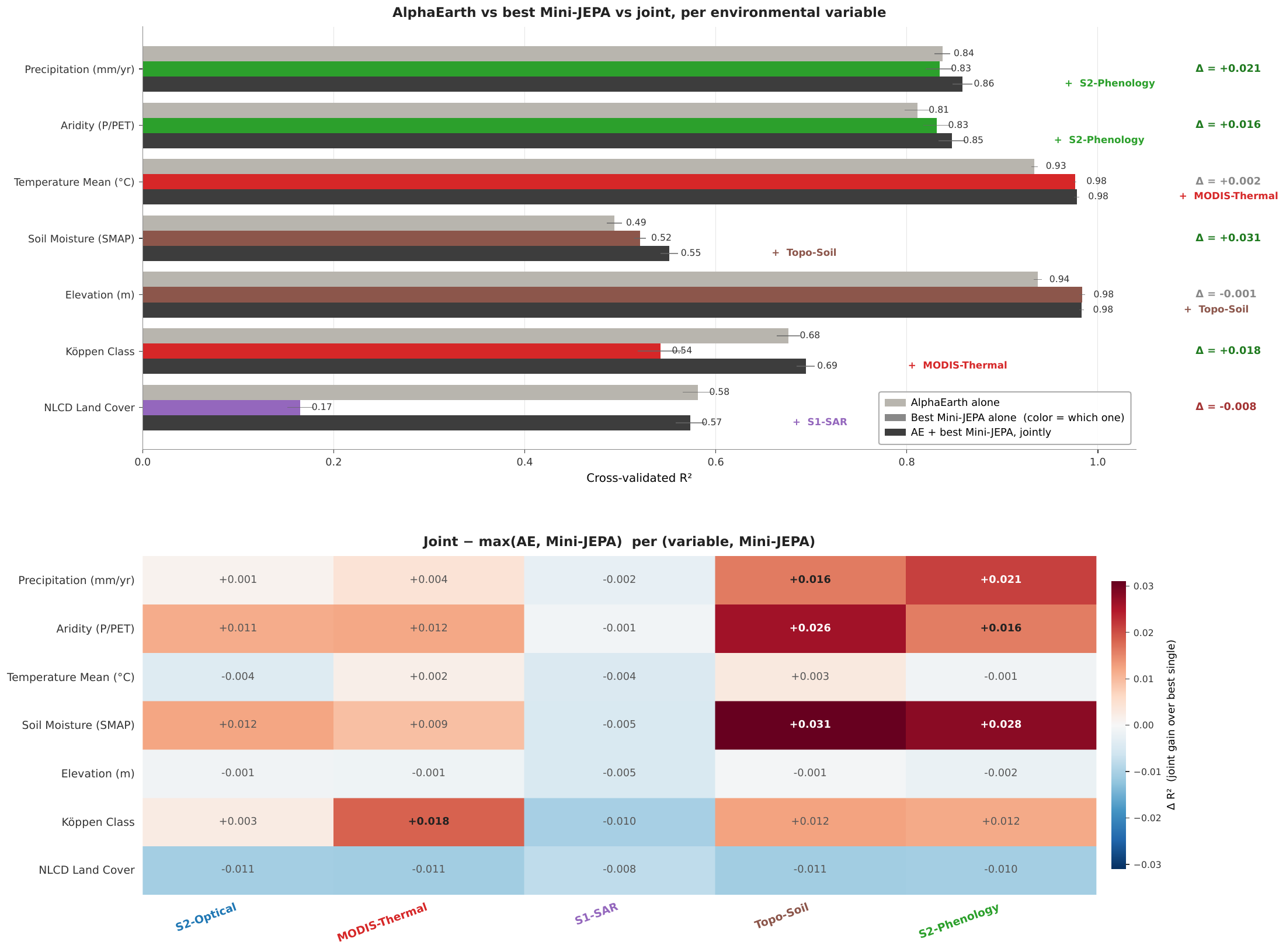}
\caption{Mini-JEPA complementarity with AlphaEarth. \textbf{Top:} per-variable bar plot of cross-validated $R^2$ for AlphaEarth alone (light grey), best Mini-JEPA alone (colored by which Mini-JEPA wins), and the joint AlphaEarth + best Mini-JEPA model (dark grey). $\Delta$ values on the right report the joint gain over the better single source. \textbf{Bottom:} $\Delta R^2$ matrix per (variable, Mini-JEPA) pair. Positive entries (red) indicate the Mini-JEPA carries information AlphaEarth does not represent for that variable.}
\label{fig:complementarity}
\end{figure}

The third question is whether each Mini-JEPA carries information that AlphaEarth does not already represent, and for which variables the complementarity is strongest. Figure~\ref{fig:complementarity} reports cross-validated $R^2$ for each of the seven environmental variables under three feature sets: AlphaEarth alone, the best Mini-JEPA alone, and the two concatenated.

For four variables, the joint feature set outperforms either source on its own. Joint Topo-Soil and AlphaEarth predict soil moisture (SMAP) with $\Delta R^2 = +0.031$ over the better single source (AlphaEarth alone at $R^2 = 0.49$, Topo-Soil alone at $R^2 = 0.52$, joint at $R^2 = 0.55$). Joint Topo-Soil and AlphaEarth predict aridity with $\Delta R^2 = +0.026$, joint S2-Phenology and AlphaEarth predict precipitation with $\Delta R^2 = +0.021$, and joint S2-Phenology and AlphaEarth predict soil moisture with $\Delta R^2 = +0.028$. The gains are small in absolute terms but consistent in sign across the hydrologically meaningful targets.

For temperature and elevation, the joint model is essentially tied with the better single source. This is unsurprising: AlphaEarth alone already reaches $R^2 = 0.93$ on temperature and $R^2 = 0.94$ on elevation, leaving little room for additional information to register at the resolution of our patch corpus. The MODIS-Thermal model reaches a comparable $R^2 = 0.97$ on temperature, and Topo-Soil reaches $R^2 = 0.97$ on elevation, but combining either with AlphaEarth produces no further improvement.

For NLCD land cover, the joint score sits below AlphaEarth alone for every Mini-JEPA. We interpret this as a sample-size and regularization effect rather than a substantive finding: NLCD is the variable on which every Mini-JEPA performs worst on its own, and concatenating a weak feature set with a stronger one introduces noise that the Random Forest cannot fully suppress at the patch counts in our corpus.

The full $\Delta R^2$ matrix (Figure~\ref{fig:complementarity}, bottom) shows that the largest positive joint gains concentrate in two columns: Topo-Soil and S2-Phenology. The other three Mini-JEPAs (S2-Optical, S1-SAR, MODIS-Thermal) generally produce small or negative joint gains. We read this as the cleanest empirical statement of which specialist embeddings are non-redundant with a planetary-scale generalist. The thermal and broad-optical signals already appear to be well-represented in AlphaEarth. Topographic, soil, and seasonal-phenology signals are not, and adding them improves AlphaEarth-based reconstructions for the variables they carry physical information about.

The complementarity pattern also speaks to the substitution case. On the variables where a single Mini-JEPA already reaches scores comparable to AlphaEarth (elevation under Topo-Soil, temperature under MODIS-Thermal, precipitation under S2-Phenology), the Mini-JEPA can stand in for AlphaEarth for that variable specifically. When AlphaEarth is unavailable or impractical to deploy, the corresponding Mini-JEPA reaches the same predictive ceiling on its matched variable with 22M parameters and a workstation-scale training budget.

\subsection{A router LLM composes the fleet into an agentic hydrologic intelligence system} \label{sec:results_agent}

The fourth question is whether a router LLM, informed by per-modality reference cards, can compose the fleet into a working agentic system, and whether the routed fleet improves LLM-judged response quality.

\subsubsection{Routing accuracy}

\begin{figure}[t]
\centering
\includegraphics[width=\linewidth]{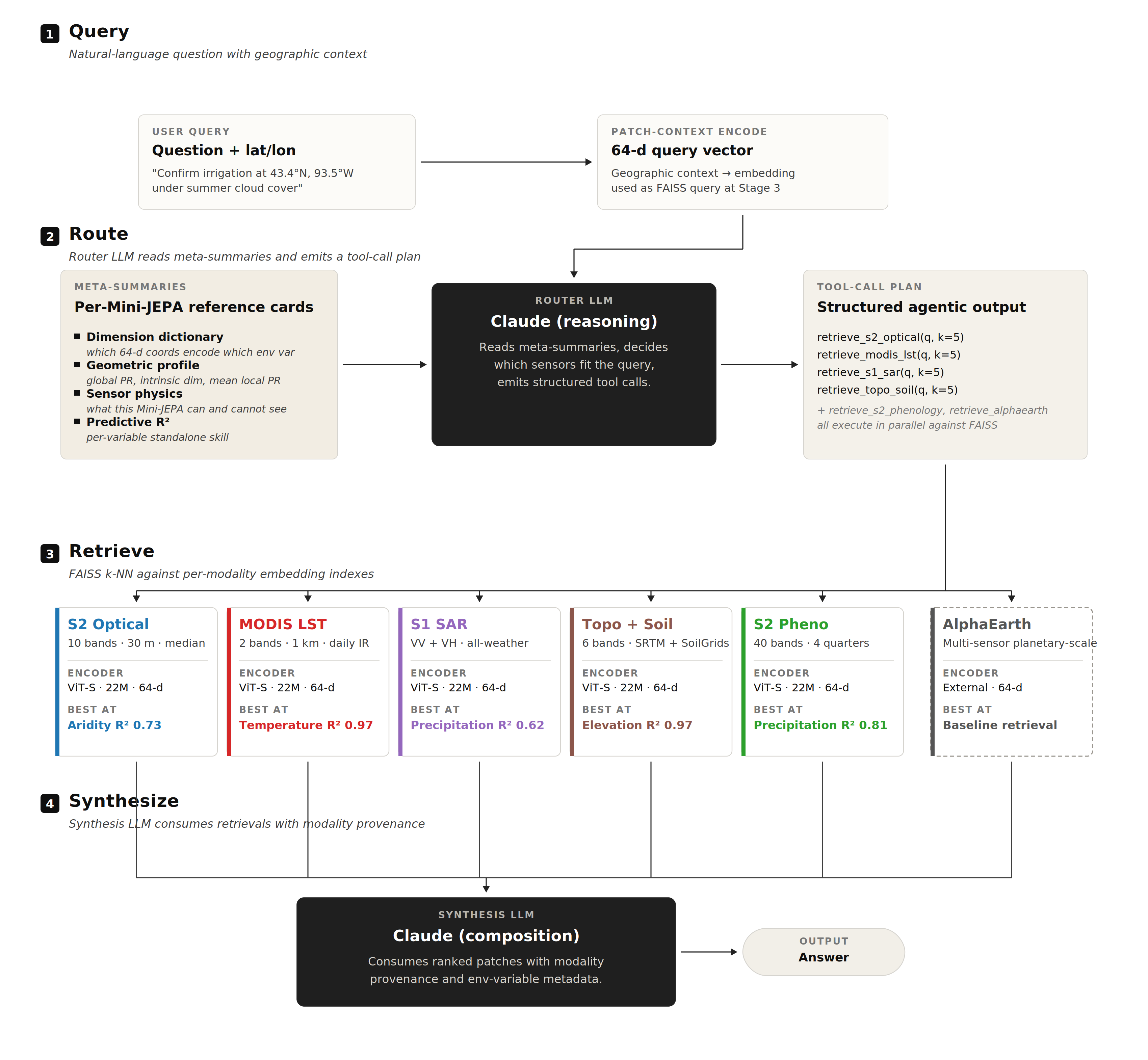}
\caption{Agentic system architecture. A user query with geographic context is encoded into a 64-dimensional query vector by each selected Mini-JEPA. The router LLM reads compact per-modality reference cards (dimension dictionary, geometric profile, sensor physics, predictive $R^2$ table) and emits a structured tool-call plan that selects any subset of the five Mini-JEPAs and optionally AlphaEarth. Each selected retrieval runs a $k$-NN search against a per-modality FAISS index. A synthesis LLM consumes the ranked retrievals with explicit modality provenance and composes the final answer.}
\label{fig:agent_architecture}
\end{figure}

The router LLM consumes the per-modality reference cards described in Section~\ref{sec:methods_agent} and emits a structured plan specifying which Mini-JEPAs to query for each question. Figure~\ref{fig:agent_architecture} shows the full agent flow, from query encoding through routing to multi-modality retrieval and synthesis. Each retrieval tool runs a $k$-nearest-neighbor search against a per-modality FAISS index over the 9{,}704 mean-pooled 64-d embeddings.

Across all question categories with a defined expected-modality set, the router selects at least one expected modality on every question (100\% expected-modality hit rate). The selection distribution across question categories shows that the router does not collapse onto a single favorite. On SAR-favorable questions, the router selects S1-SAR most often, with Topo-Soil as a frequent companion. On multi-modality questions, S1-SAR and S2-Optical co-occur, often supplemented by Topo-Soil. On AE-favorable questions, the router still selects from the Mini-JEPA fleet (S2-Phenology and S2-Optical) alongside AlphaEarth when dual-RAG is available. The routing distribution therefore reflects the question's physical content rather than a fixed prior.

\subsubsection{Response quality under three retrieval conditions}

\begin{figure}[t]
\centering
\includegraphics[width=\linewidth]{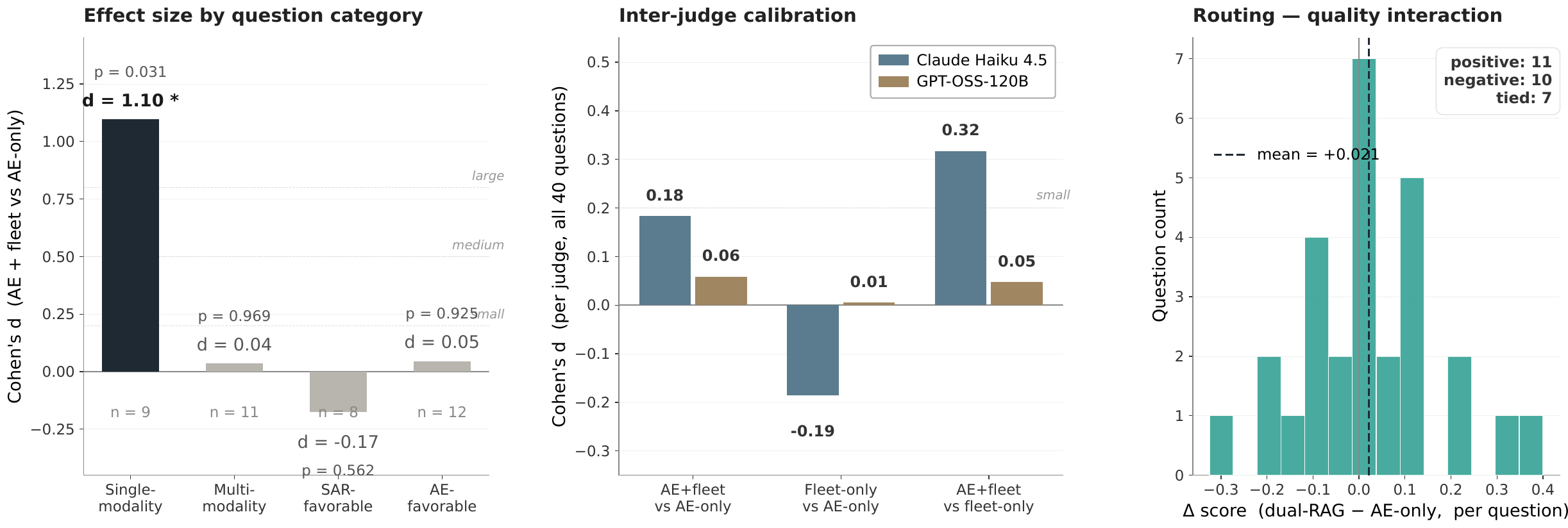}
\caption{Agent evaluation across three retrieval conditions and four question categories. \textbf{Left:} effect size (Cohen's $d$) of AE + Fleet vs.~AE-only, broken down by question category. The single-modality category shows a large positive effect ($d = 1.10$, $p = 0.031$, $n = 9$). \textbf{Middle:} inter-judge calibration showing per-judge Cohen's $d$ across the three condition contrasts (AE + Fleet vs.~AE-only, Fleet-only vs.~AE-only, AE + Fleet vs.~Fleet-only). \textbf{Right:} per-question $\Delta$ score distribution (AE + Fleet minus AE-only) across all 40 questions; mean $\Delta = +0.021$, with 11 questions improving, 10 declining, and 7 tied.}
\label{fig:experimental_results}
\end{figure}

For each of 40 hydrologic questions stratified into four categories, we generated responses under three conditions: AE only (AlphaEarth retrieval, no fleet), Fleet only (router selects from the five Mini-JEPAs, AlphaEarth disabled), and AE + Fleet (router selects from all six, the full dual-RAG setup). Each response was scored independently by two LLM judges. Figure~\ref{fig:experimental_results} reports the comparison.

The aggregate weighted scores across all 40 questions sit close to each other on the 1-to-5 scale (mean of approximately 4.4-4.5 across the three conditions). The fleet-based conditions do not improve aggregate response quality relative to AE-only at the resolution of the full question set. The mean per-question gain of AE + Fleet over AE-only is $\Delta = +0.021$ across all 40 questions (Figure~\ref{fig:experimental_results}, right), with 11 questions improving, 10 declining, and 7 tied.

The category breakdown tells a different story (Figure~\ref{fig:experimental_results}, left). On \emph{single-modality} questions, AE + Fleet outperforms AE-only with a large effect size (Cohen's $d = 1.10$, $p = 0.031$, $n = 9$). These are the questions whose physical signal sits cleanly in one sensor: the kind of question for which a sensor-specialized embedding is most directly useful. The effect size is positive and small to medium on the other three categories (\emph{multi-modality}: $d = 0.04$; \emph{SAR-favorable}: $d = -0.17$; \emph{AE-favorable}: $d = 0.05$), none of which reach significance.

The reason the aggregate effect is modest while the single-modality effect is large is, in our reading, that the LLM-as-Judge scores saturate near the top of the 1-to-5 scale for the strong language models used as the system LLM (Claude Opus 4.7 and Sonnet 4.6). On broad characterization questions, both AE-only and AE + Fleet produce responses the judges score above 4.5 regardless of retrieval source, leaving little headroom for retrieval-quality differences to register. On physics-matched single-modality questions, the headroom is recovered: the routed fleet's targeted retrieval produces responses with grounding and scientific accuracy that AE-only retrieval cannot match, and the judges register the gap.

Inter-judge calibration (Figure~\ref{fig:experimental_results}, middle) shows that the two judges agree on the sign of the effect for AE + Fleet vs.~AE-only (Haiku 4.5: $d = 0.18$; GPT-OSS-120B: $d = 0.06$). The Fleet-only condition produces a slightly negative effect relative to AE-only under Haiku ($d = -0.19$), suggesting that AlphaEarth's broad-coverage retrievals add value even on questions the fleet alone can address. The AE + Fleet condition is therefore the operationally interesting one: it preserves AlphaEarth's broad-context retrieval and adds the specialist information from the fleet when the question calls for it.

The two operational claims that follow from this evaluation are: (1) on questions whose physical signal matches a single sensor, dual retrieval over AlphaEarth and the routed fleet outperforms AlphaEarth alone with a large effect size; and (2) on the same physics-matched questions, the routed fleet by itself reaches scores comparable to AlphaEarth, suggesting Mini-JEPAs can substitute for a planetary-scale generalist when one is unavailable, restricted, or impractical to deploy.

\section{Conclusion} \label{sec:conclusion}

We introduced a fleet of five Mini-JEPA foundation models, each pretrained on a different Google Earth Engine satellite product under an identical Vision Transformer architecture and an identical JEPA training recipe. Holding the architecture fixed across the fleet meant that the only systematic difference between models was the sensor each one saw during pretraining. We characterized each Mini-JEPA along three axes that mirror our prior studies on AlphaEarth \citep{rahman2026physically, rahman2026characterizing}, covering dimension-level physical interpretability, manifold geometry, and complementarity with a planetary-scale generalist embedding. We then built a routing agent that reads compact per-modality reference cards and decides which Mini-JEPAs to query for a given hydrologic question, with a synthesis layer that composes the resulting retrievals into a final answer. The full system was evaluated with paired LLM-as-Judge scoring across four question categories.

The findings point to four broad implications. First, sensor physics shapes specialization. Each Mini-JEPA's strongest predictive skill matches a variable its sensor physically observes. Some variables (precipitation in particular) are well-predicted by more than one sensor through different physical pathways, but the diagonal pattern of best-matched skill is consistent across the fleet. Second, the five Mini-JEPAs do not share a manifold structure: their global effective dimensionalities and local geometric complexities differ in ways that reflect what each sensor sees. Third, Mini-JEPAs trained on topography-and-soil and on multi-temporal phenology carry information that a planetary-scale generalist does not already represent, particularly for soil moisture, aridity, and precipitation. Fourth, an LLM router informed by per-modality reference cards reliably picks the physically appropriate sensors for a question, and on questions whose signal matches a single sensor, dual retrieval over the routed fleet and the generalist outperforms the generalist alone with a large effect size. The same physics-matched comparison also suggests Mini-JEPAs can substitute for a generalist when one is unavailable, restricted, or impractical to deploy.

Several limitations qualify these findings. The patch corpus is restricted to CONUS and to a single calendar year. The Mini-JEPAs we report should not be expected to transfer cleanly to other climates, biomes, or years without retraining or adaptation. The LLM-as-Judge evaluation, while paired across two judges and grounded in a five-rubric scoring protocol, saturates near the top of the 1-to-5 scale for the strong language models used as the system LLM. The aggregate response-quality comparison is therefore lower-resolution than the per-category breakdown, which is where the largest effects appear. The question set itself contains 40 questions across four categories, which is enough to detect a large effect in the single-modality subset but not enough to resolve smaller effects in the others. The fleet currently has five members chosen along the sensor-physics axis; specialization along hydrologic regime (snow, arid, urban, tropical) or temporal scale (event, seasonal, interannual) is left to future work. Finally, the LLM-as-Judge scoring inherits whatever calibration and bias each judge brings to environmental reasoning, and judge agreement between Claude Haiku 4.5 and GPT-OSS-120B is positive but not perfect.

Several directions extend the work naturally. Specialization along hydrologic regime rather than along sensor product would test whether the fleet design generalizes from physical-sensor specialization to process-regime specialization, where each Mini-JEPA captures the dynamics of one hydrologic state of the surface. Extending the corpus beyond CONUS and beyond a single year would test temporal and geographic generalization, both of which are open questions for any small specialist foundation model. The agent itself can be made richer: routing currently selects which sensors to query, but future versions could decide how to combine retrievals, query at multiple spatial scales, or invoke domain-specific scientific tools alongside retrieval. 

The broader implication of the fleet-plus-router design is methodological. Building useful foundation models for environmental science has converged on a recipe that requires planetary-scale compute, planetary-scale data pipelines, and access to expensive compute systems. The pattern we report here suggests a complementary path: small specialist models that can be trained, audited, and extended on commodity hardware (a personal computer with a GPU), composed by an agent that reads each model's own reference card before deciding which to consult. For research groups and institutions that cannot easily access planetary-scale generalists, the fleet design lowers the barrier to building bespoke environmental embeddings that fit their own questions. For any group, the fleet design produces systems that are easier to inspect and easier to correct than a single closed model, which matters for environmental decisions whose downstream consequences are concrete. The work points to a future in which generalist embeddings and specialist fleets coexist, with the agent selecting between them when the question's physics calls for one rather than the other.

\bibliographystyle{elsarticle-num-names} 
\bibliography{bibliography.bib}

\section*{Acknowledgements}

The author thanks the Dartmouth Libraries for institutional support throughout this work. We acknowledge Google Earth Engine \citep{gorelick2017gee} for providing access to the satellite imagery and environmental data products used to build the Mini-JEPA training corpora. All Mini-JEPA pretraining and per-modality evaluation was performed on a single NVIDIA RTX 5090 workstation, which made the fleet design tractable on commodity hardware.

\section*{Data and Code Availability}

All scripts for data acquisition, Mini-JEPA pretraining, per-modality evaluation, agent routing, and LLM-as-Judge scoring, together with the trained Mini-JEPA checkpoints, the 9{,}704-patch corpus with co-located environmental labels, per-modality FAISS indexes, and the agent evaluation outputs (responses, judge scores, routing logs), are archived on Zenodo at \href{https://doi.org/10.5281/zenodo.20170560}{10.5281/zenodo.20170560}. Source satellite products were retrieved from Google Earth Engine and are listed with their dataset IDs in Table~\ref{tab:datasets}. The AlphaEarth Foundation V1 annual embeddings \citep{brown2025alphaearth} are publicly available through the \texttt{GOOGLE/SATELLITE\_EMBEDDING/V1/ANNUAL} ImageCollection on Google Earth Engine.

\section*{Declaration of Generative AI Use}

Large language models were used in this work both as research instruments and as a coding assistant. Claude Sonnet 4.6 and Claude Opus 4.7 served as the routing and synthesis LLMs in the agentic system evaluated in Section~\ref{sec:results_agent}. Claude Haiku 4.5 and GPT-OSS-120B served as the two independent LLM judges in the response-quality evaluation. Claude Code was used to assist with software development for the data acquisition, pretraining, evaluation, and agent pipelines. 

\end{document}